\documentclass{bmvc2k}

\usepackage{mathtools}
\usepackage{amsfonts}
\usepackage{pifont}
\usepackage{booktabs}
\usepackage{flexisym}

\title{Construct Dynamic Graphs for Hand Gesture Recognition via Spatial-Temporal Attention}

\addauthor{Yuxiao Chen}{yc984@cs.rutgers.edu}{1}
\addauthor{Long Zhao}{lz311@cs.rutgers.edu}{1}
\addauthor{Xi Peng}{xipeng@udel.com}{2}
\addauthor{Jianbo Yuan}{jyuan10@cs.rochester.edu}{3}
\addauthor{Dimitris N. Metaxas}{dnm@cs.rutgers.edu}{1}

\addinstitution{
 Department of Computer Science\\
 Rutgers University\\
 New Jersey, USA
}
\addinstitution{
 Department of Computer and Information Sciences\\
 University of Delaware\\
 Delaware, USA
}
\addinstitution{
 Department of Computer Science\\
 University of Rochester\\
 New York, USA
}

\runninghead{Yuxiao Chen}{Dynamic Graphs for Hand Gesture Recognition}

\def\eg{\emph{e.g}\bmvaOneDot}
\def\ie{\emph{i.e}\bmvaOneDot}

\def\etal{\emph{et al}\bmvaOneDot}

\begin{document}
\maketitle

\begin{abstract}
We propose a Dynamic Graph-Based Spatial-Temporal Attention (DG-STA) method for hand gesture recognition. The key idea is to first construct a fully-connected graph from a hand skeleton, where the node features and edges are then automatically learned via a self-attention mechanism that performs in both spatial and temporal domains. 
We further propose to leverage the spatial-temporal cues of joint positions to guarantee robust recognition in challenging conditions. 
In addition, a novel spatial-temporal mask is applied to significantly cut down the computational cost by 99\%. We carry out extensive experiments on benchmarks (DHG-14/28 and SHREC'17) and prove the superior performance of our method compared with the state-of-the-art methods. The source code can be found
at \url{https://github.com/yuxiaochen1103/DG-STA}.
\end{abstract}

\section{Introduction}
\label{sec:intro}

Hand gesture recognition has been an active research area due to its wide range of applications such as human computer interaction, gaming and nonverbal communication analysis including sign language recognition~\cite{nikam2016sign,wachs2011vision,rautaray2015vision}. Previous work can be classified into two categories based on the modality of their inputs: image-based~\cite{freeman1995orientation,molchanov2015hand,wang2015superpixel} and skeleton-based~\cite{de2016skeleton,hou2018spatial,de2017shrec,chen2017motion,nunez2018convolutional} methods. Image-based methods take RGB or RGB-D images as inputs and rely on image-level features for recognition. On the other hand, skeleton-based methods make predictions by a sequence of hand joints with 2D or 3D coordinates. They are more robust to varying lighting conditions and occlusions given the accurate joint coordinates. Thanks to the low-cost depth cameras (\eg, Microsoft Kinect or Intel RealSense) and great progress made on hand pose estimation~\cite{oberweger2015hands,oberweger2017deepprior,oberweger2015training}, accurate coordinates of hand joints are easy to be obtained. Therefore, we follow the skeleton-based method in this work.

Conventional methods~\cite{ohn2013joint,de2016skeleton,de2017shrec} of skeleton-based hand gesture recognition aim to design powerful feature descriptors to model the action of hands. However, these hand-crafted features have limited generalization capability. Recent studies~\cite{hou2018spatial,chen2017motion,nunez2018convolutional} have achieved significant improvement using deep learning. They usually concatenate the joint coordinates into a tensor which is fed into a neural network, and then hand features are directly learned by the network during training. Nevertheless, the spatial structures and temporal dynamics of hand skeletons are not explicitly exploited in these deep learning based approaches.

More recent studies~\cite{yan2018spatial,si2018skeleton,zhao2019semantic} attempt to incorporate structures and dynamics of skeletons based on skeleton graphs. Specifically, given a sequence of skeletons, they define a spatial-temporal graph where structures and dynamics of skeletons are embedded. The feature representation of the graph is then extracted for action recognition. 
However, a pre-defined graph with fixed structure lacks the flexibility to capture the variance and dynamics across different actions, yielding sub-optimal performance in practice.  

\begin{figure}[t]
\includegraphics[width=\textwidth]{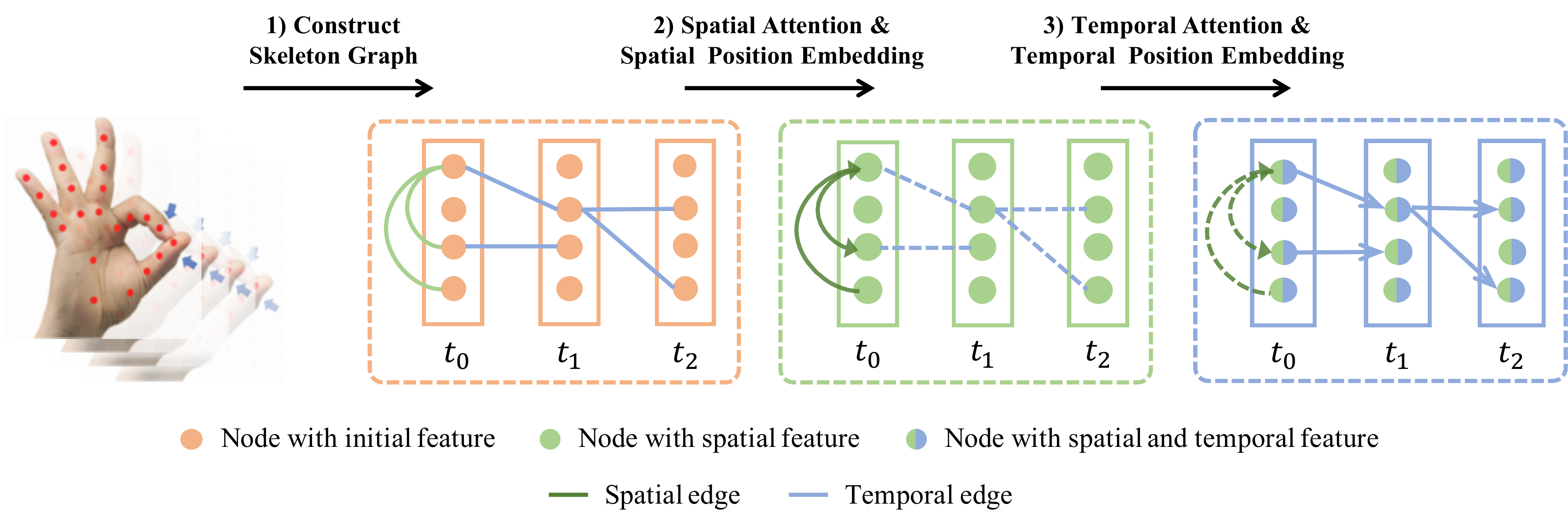}
\caption{Illustration of our method. The nodes in the graph correspond to hand joints and the dashed lines represent disconnected edges. The proposed DG-STA calculates edge weights and learn node features in both spatial and temporal domains of the hand skeleton graph.}
\label{fig:overview}
\end{figure}

To this end, we propose a \emph{Dynamic Graph-Based Spatial Temporal Attention (DG-STA)} model for hand gesture recognition. The key idea is to perform a self-attention mechanism in both spatial and temporal domains to modify a unified graph dynamically in order to model different actions.
Figure~\ref{fig:overview} gives an overview of our approach. 
There are three crucial designs that distinguish our approach from previous methods.
\textbf{First}, instead of a pre-defined graph with fixed structure, we propose to construct a unified graph where the edges and nodes are dynamically optimized according to different actions. This makes it is possible to achieve action-specific graphs with improved expressive power.
\textbf{Second}, we propose \emph{spatial-temporal position embedding} which improves the temporal position embedding~\cite{vaswani2017attention}. It encodes the identity and temporal order information of each node in the graph. Combining node features with their position embeddings can further improve the performance of our approach.
\textbf{Third}, to implement our DG-STA more efficiently, we present a novel \emph{spatial-temporal mask operation} which is directly applied to the matrix of scaled dot-products among all nodes. It significantly improves the computational efficiency of our model and allows easier input data arrangement.


To evaluate the effectiveness of our approach, we conduct comprehensive experiments on two standard benchmarks: DHG-14/28 Dataset~\cite{de2016skeleton} and SHREC'17 Track Dataset~\cite{de2017shrec}. The results demonstrate that our method outperforms the state-of-the-art methods. In summary, our main contributions are summarised as follows:
\begin{itemize}
\item We propose Dynamic Graph-Based Spatial-Temporal Attention (DG-STA) for skeleton-based hand gesture recognition. The structures and dynamics of hand skeletons can be learned automatically and more efficiently by our approach.
\item We propose spatial-temporal position embedding which encodes the identity and temporal order information of nodes to boost the performance of our model, and a spatial-temporal mask operation for efficient implementation of DG-STA.
\item We conduct comprehensive experiments to validate our approach on two standard benchmarks. The proposed DG-STA achieves the state-of-the-art performance.
\end{itemize}
\section{Related Work}

In this section, we review recent work on self-attention  and recent developments in skeleton-based action recognition, which motivated our approach.

\textbf{Self-Attention.} 
The self-attention mechanism is widely used in computer vision and natural language processing tasks~\cite{chen2018factual,chen2018twitter,chen2018commerce,ZhangGMO19,lin2017structured,tan2018deep,verga2018simultaneously,tian2018cr}.
Vaswani~\etal~\cite{vaswani2017attention} proposed to apply the self-attention  module to model temporal and semantic relationships among words within a sentence for machine translation. Instead, we study applying the self-attention mechanism to learn spatial-temporal information contained in hand skeletons represented by graphs, which are largely different from sequences. Graph Attention Networks (GATs)~\cite{velivckovic2017graph} employed self-attention to learn node embeddings of graphs. By contrast, our approach is able to capture additional temporal information as well as node identities. 


\textbf{Skeleton-Based Hand Gesture Recognition.} Skeleton-based hand gesture is a well-studied but still challenging task. Traditional methods~\cite{ohn2013joint,de2016skeleton,de2017shrec,lu2003using} mainly focus on designing powerful hand feature descriptors. Smedt~\etal~\cite{de2016skeleton} proposed the Shape of Connected Joints descriptor to represent the hand shape of hand skeleton. Recent studies~\cite{hou2018spatial,chen2017motion,nunez2018convolutional} apply deep neural networks for this task and achieve significant performance improvement. Convolutional Neural Networks and Long Short-Term Memory are leveraged to learn the spatial and temporal features from the sequence of hand joints for hand gesture classification in~\cite{nunez2018convolutional}. One limitation of these learning-based methods is that they do not explicitly explore the structures and dynamics of human hands.

\textbf{Skeleton-Based Human Action Recognition.} Recent studies in skeleton-base human action recognition~\cite{zhao2018learning,tang2018quantized,peng2018jointly} started to incorporate structures and dynamics of human bodies by building skeleton graphs~\cite{yan2018spatial,si2018skeleton,zhao2019semantic}. This idea is first introduced by~\cite{dlid2017graph} which employs Graph CNNs. Recently, Yan~\etal~\cite{yan2018spatial} built a skeleton graph based on the natural structure of human body, and extract its representation by Graph Convolution Networks~\cite{kipf2016semi} for action recognition. Nevertheless, it is difficult to define an optimal skeleton graph which represents all action-specific structures and dynamics information. Instead, our method can automatically learn multiple action-specific graphs with the multi-head attention mechanism~\cite{vaswani2017attention}, which efficiently encode structures and dynamics of hand gestures.

\section{Methodology}

The overview of our approach is shown in Figure~\ref{fig:overview}. First, a fully-connected skeleton graph is constructed from the input sequence of hand skeletons as described in Section~\ref{sec:init_g}. In Section~\ref{sec:sta_graph}, we devise DG-STA to learn the edge weights and node embeddings within the graph. The learned node features by DG-STA are then average-pooled into a vector which captures the structures and dynamics of the input skeleton graph. We use it for hand gesture classification.
Section~\ref{sec:pos_embedding} presents the spatial-temporal position embedding which is combined with node features to incorporate node identity and temporal order information contained in the hand skeletons. Moreover, a spatial-temporal mask operation is introduced in Section~\ref{sec:efficient_imp} which implements our proposed DG-STA more efficiently.     

\subsection{Skeleton Graph Initialization}
\label{sec:init_g}
Given a video of $T$ frames, $N$ hand joints are extracted from each frame to represent the hand skeleton. Then a fully-connected skeleton graph $G = (V, E)$ is constructed from this sequence of hand skeletons. Let $V = \{v_{(t, i)}|t = 1, \dots , T, i = 1, \dots  , N\}$ denote the node set where $v_{(t, i)}$ represents the $i$-th hand joint at the time step $t$. The node features are represented by $F = \{\textbf{f}_{(t,i)}|t = 1, \dots , T, i = 1, \dots  , N\}$, where $\textbf{f}_{(t,i)}$ indicates the feature vector of the node $v_{(t,i)}$. They are extracted from the 3D coordinates of nodes. Note that each node is connected with all other nodes including itself. For clarity, we define three types of edges on the edge set $E$ as follows. 
\begin{itemize}
\item A spatial edge $v_{(t,i)} \rightarrow v_{(t,j)}(i \neq j)$ connects two different nodes at the same time step.
\item A temporal edge $v_{(t,i)} \rightarrow v_{(k,j)}(t \neq k)$ connects two nodes at different time steps.
\item A self-connected edge $v_{(t,i)} \rightarrow v_{(t,i)}$ connects the node with itself. 
\end{itemize}


%

\subsection{Dynamic Graph Construction via Spatial-Temporal Attention}
\label{sec:sta_graph}


The proposed  DG-STA consists of the spatial attention model $\mathbf{A}_S$ and temporal attention model $\mathbf{A}_T$ which are employed to extract spatial and temporal information from the hand skeleton graph respectively. Both $\mathbf{A}_S$ and $\mathbf{A}_T$ are based on multi-head attention~\cite{vaswani2017attention}. $\mathbf{A}_S$ first takes the initial node features $F$ as the input and updates them to encode spatial information. The updated node features are then fed to $\mathbf{A}_T$ to further learn temporal information. Finally, the results are average-pooled to a vector which is used as the feature representation of the skeleton graph for classification.

Specifically, given the input feature $\mathbf{f}_{(t,i)}$ of the node $v_{(t,i)}$ in the skeleton graph, the $h$-th spatial attention head first applies three fully-connected layers to map $\mathbf{f}_{(t,i)}$ into the key, query and value vectors respectively, which are formulated as:
\begin{equation}
	\label{eq:Q_K_V}
        \mathbf{K}_{(t,i)}^h = W_{K}^h\mathbf{f}_{(t,i)},\  \quad
        \mathbf{Q}_{(t,i)}^h = W_{Q}^h\mathbf{f}_{(t,i)},\ \quad
        \mathbf{V}_{(t,i)}^h = W_{V}^h\mathbf{f}_{(t,i)},\
\end{equation}
where $\mathbf{K}_{(t,i)}^h$, $\mathbf{Q}_{(t,i)}^h$ and $\mathbf{V}_{(t,i)}^h$ represent the key, query and value vectors of the node; $\mathbf{W}_K^h$, $\mathbf{W}_Q^h$ and $\mathbf{W}_V^h$ are the corresponding weight matrices of the three fully-connected layers of the $h$-th spatial attention head.

The spatial attention head computes the weights of the spatial and self-connected edges in two steps. First, it calculates the ``scaled dot-product''~\cite{vaswani2017attention} between the query vectors and key vectors of the nodes within the same time step. Then it normalizes the results by a Softmax function. These two steps are formulated as: 
\begin{equation}
	\label{eq:s_att}
        u_{(t,i) \rightarrow (t,j)}^h =\frac{{\langle\ \mathbf{Q}_{(t,i)}^h,\mathbf{K}_{(t,j)}^h \rangle}}{\sqrt{d}},\ \quad
        \alpha_{(t,i) \rightarrow (t,j)} ^ h =\frac{\exp \left(u_{(t,i) \rightarrow (t,j)} ^ h\right)}{\sum_{n=1}^{N}\exp \left(u_{(t,i) \rightarrow  (t,n)}^h\right)},\
\end{equation}
where $d$ is the dimension of the key, query and value vectors; $u_{(t,i) \rightarrow  (t,j)}^h$ is the scaled dot-product of the node $v_{(t,i)}$ and $v_{(t,j)}$; $\langle\cdot,\cdot\rangle$ represents the inner product operation; $\alpha_{(t,i) \rightarrow (t,j)} ^ h$ is the attention weight between the node $v_{(t,i)}$ and $v_{(t,j)}$, which measures the importance of information from node $v_{(t,j)}$ to node $v_{(t,i)}$. Meanwhile, the weights for all temporal edges are set to 0 in order to block the information passing in the temporal domain. As a result, each spatial attention head produces a weighted skeleton graph which represents a specific type of spatial structure of the hand. 

The attention head calculates the spatial attention feature of the node $v_{(t,i)}$ as the weighted sum of the value vectors within the same time step, which is defined as: 
\begin{equation}
    \label{eq:s_value}
        \bar{\mathbf{f}}_{(t,i)}^ h =\sum_{j=1}^{N} \left(\alpha_{(t,i) \rightarrow (t,j)} ^ h\cdot \mathbf{V}_{(t,j)}^h \right),\
\end{equation}
where $\bar{\textbf{f}}_{(t,i)}^h$ denotes the spatial attention feature of the node $v_{(t,i)}$. Intuitively, the computation mechanism of the spatial attention features is essentially the process that each node in the graph sends some information to the others within the same time step and then aggregates the received information based on the learned edge weights.

The spatial attention model $\mathbf{A}_S$ finally concatenates the spatial attention features learned by all spatial attention heads into $\tilde{\mathbf{f}}_{(t,i)}$ which is employed as the spatial feature of the node $v_{(t,i)}$:
\begin{equation}
\label{eq:mutil}
\tilde{\textbf{f}}_{(t,i)} = \text{Concate}\left[\bar{\textbf{f}}_{(t,i)}^{1}, \bar{\textbf{f}}_{(t,i)}^{2}, ..., \bar{\textbf{f}}_{(t,i)}^{H}\right],
\end{equation}
where $H$ is the number of spatial attention heads. 
The obtained node features encode multiple types of structural information represented by the weighted skeleton graphs which are learned by different spatial attention heads. 

The temporal attention model $\mathbf{A}_T$ takes the output node features from the spatial attention model as the input, and then applies the above multi-head attention mechanism in the temporal domain. The node feature which is the output from the temporal attention model encodes both spatial and temporal information carried by the input sequence of the hand skeletons. We average-pool these node features to a vector as the feature representation of the input sequence for hand gesture recognition.

\subsection{Spatial-Temporal Position Embedding}
\label{sec:pos_embedding}


The original node features $F$ extracted from the coordinates of the input hand skeletons do not contain spatial identity information describing which hand joint a node corresponds to, and temporal information indicating which time step a node is at. To incorporate these messages, we propose the spatial-temporal position embedding.

Specifically, our spatial-temporal position embedding is made up of  the spatial position embedding and the temporal position embedding. The spatial one consists of $N$ vectors and each  represents a hand joint. Meanwhile, the temporal one is composed of $N \times T$ distinct vectors and each of them corresponds to a node in the hand skeleton graph. The feature vector of a specific node is added with the corresponding spatial and temporal position embedding vectors before fed into the DG-STA. Therefore, we have: 
\begin{equation}
\hat{\mathbf{f}}_{(t,i)} = \mathbf{A}_T\left(\mathbf{p}_{(t,i)}^T + \mathbf{A}_S\left(\mathbf{f}_{(t,i)} + \mathbf{p}_{(i)}^S\right)\right),
\end{equation}
where $\hat{\mathbf{f}}_{(t,i)}$ denotes the final output feature of node $v_{(t,i)}$, $\mathbf{p}_{(i)}^S$ is the spatial position embedding of the $i$-th hand joint, and $\mathbf{p}_{(t,i)}^T$ denotes the temporal position embedding of the $i$-th hand joint at $t$-th time step. These embeddings are with the same dimension as $\textbf{f}_{(t,i)}$, and their values are set using the sine and cosine functions of different frequencies following~\cite{vaswani2017attention}.

\subsection{Efficient Implementation}
\label{sec:efficient_imp}

 It is not straightforward to implement the proposed DG-STA because the input data have to be arranged in a complex format. However, we find that the computation of attention weights and features without domain constraints is straightforward, which can be implemented efficiently using matrix multiplication operations. Therefore, we propose a novel scheme to facilitate the implementation of the DG-STA. The main idea is to first compute the matrix of the scaled dot-products among all nodes and then apply the proposed spatial-temporal mask operation to the matrix in order to let the model focus on the spatial or temporal domain. 
 
\begin{figure}[h]
\includegraphics[width=1\textwidth]{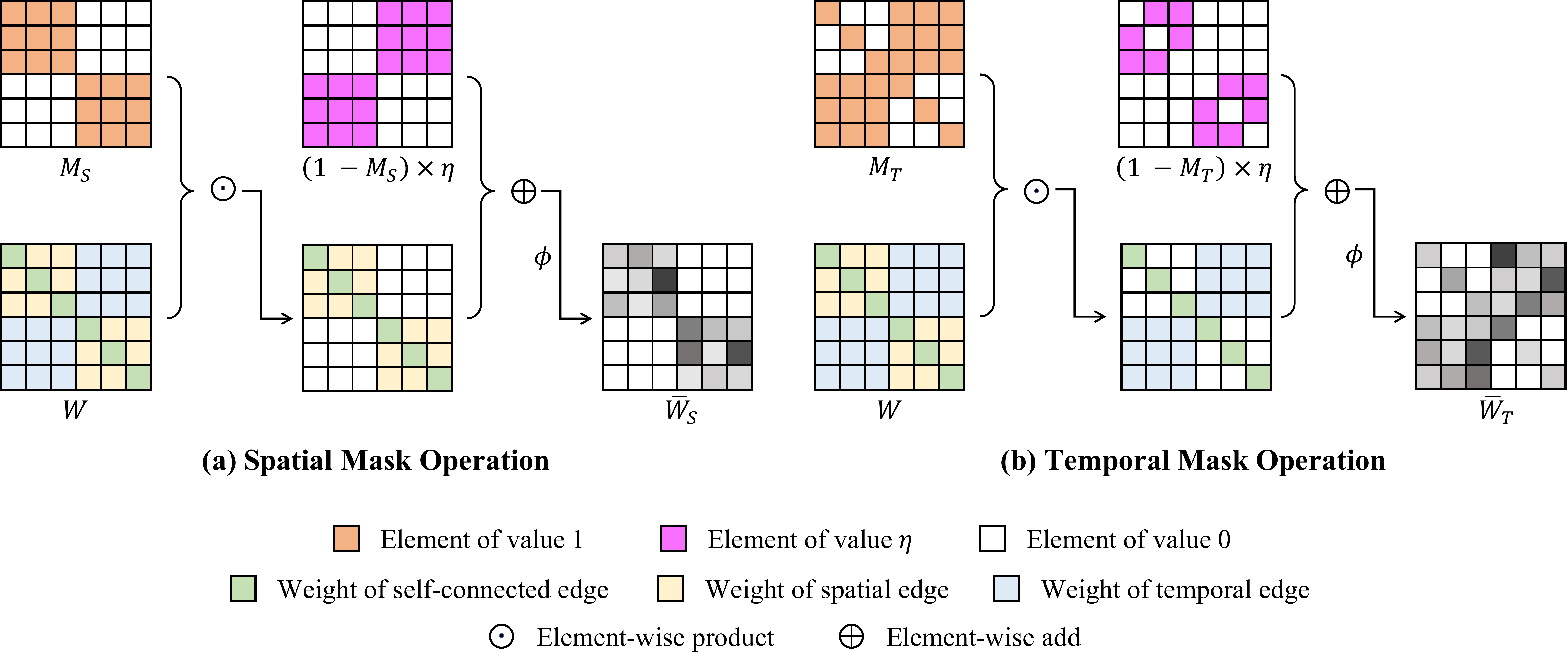}
\caption{Illustration of the proposed spatial and temporal mask operations.}
\label{fig:mask_op}
\end{figure}
 
An illustration of our mask operation is shown in Figure~\ref{fig:mask_op}. For a specific attention head, we compute a query matrix $\mathbf{Q}$ where each row represents the query vector of each node, and a key matrix $\mathbf{K}$ where each row corresponds to the key vector of each node. The matrix of the scaled dot-products $\mathbf{W}$ (\ie, the edge weights before normalization) can be obtained by:
\begin{equation}
\label{eq:scale_dot}
\mathbf{W} =\mathbf{Q} \otimes \mathbf{K}^{\top},
\end{equation}
where $\otimes$ is the matrix multiplication, and $\top$ denotes the matrix transpose operation. Then the proposed spatial mask operation sets the value of each element in $\mathbf{W}$ which represents the temporal edge to $\eta$ (\ie, a number close to negative infinity) and keeps the values of other elements unchanged. Therefore, the resulting matrix after the spatial mask operation  $\bar{\textbf{W}}_S$ is calculated:
\begin{equation}
\label{eq:mask}
\bar{\mathbf{W}}_S = \phi\left(\mathbf{W} \odot \mathbf{M}_S + \left(1 - \mathbf{M}_S\right) \times \eta\right),\\
\end{equation}
where $\odot$ denotes the element-wise dot operation, $\mathbf{M}_S$ is the spatial mask where the elements are 1 if they represent the spatial or self-connect edges and 0 otherwise. We set $\eta$ to $-9\times10^{15}$ in our implementation. The Softmax activation $\phi$ essentially normalizes weights across spatial edges, because the exponential value of the $\eta$ is close to 0. As a result, all weights of the temporal edges in $\bar{\mathbf{W}}_S$ are set to 0. Equations~(\ref{eq:scale_dot}) and (\ref{eq:mask}) efficiently implement the calculation of edge weights in the spatial domain formulated in Equation~(\ref{eq:s_att}). Moreover, the matrix $\bar{\mathbf{W}}_S$ can be directly employed to implement the computation of node features represented by Equation~(\ref{eq:s_value}) by performing the matrix multiplication operation with the matrix of value vectors. 

We define the temporal mask operation following the same way as Equation~(\ref{eq:mask}). The difference is that we use the temporal mask $\mathbf{M}_T$ instead of $\mathbf{M}_S$ to compute the matrix after the temporal mask operation  $\bar{\mathbf{W}}_T$. The elements of $\mathbf{M}_T$ are 1 if they represent the temporal or self-connect edges and 0 otherwise. With the help of the proposed spatial-temporal mask operation, we experimentally find that the computation time is reduced by 99\%.
\section{Experiments}
In this section, we first describe our network structure in Section~\ref{sec:arch}. In Section~\ref{sec:settings}, we introduce the datasets and settings employed in the experiments. Then we conduct ablation studies in Section~\ref{sec:abl_study} to evaluate the effectiveness of each component proposed in our method. Finally, we report our results and comparisons with the state of the art in Section~\ref{sec:comparison}.

\subsection{Implementation Details}
\label{sec:arch}

\begin{figure}[h]
\includegraphics[width=1\textwidth]{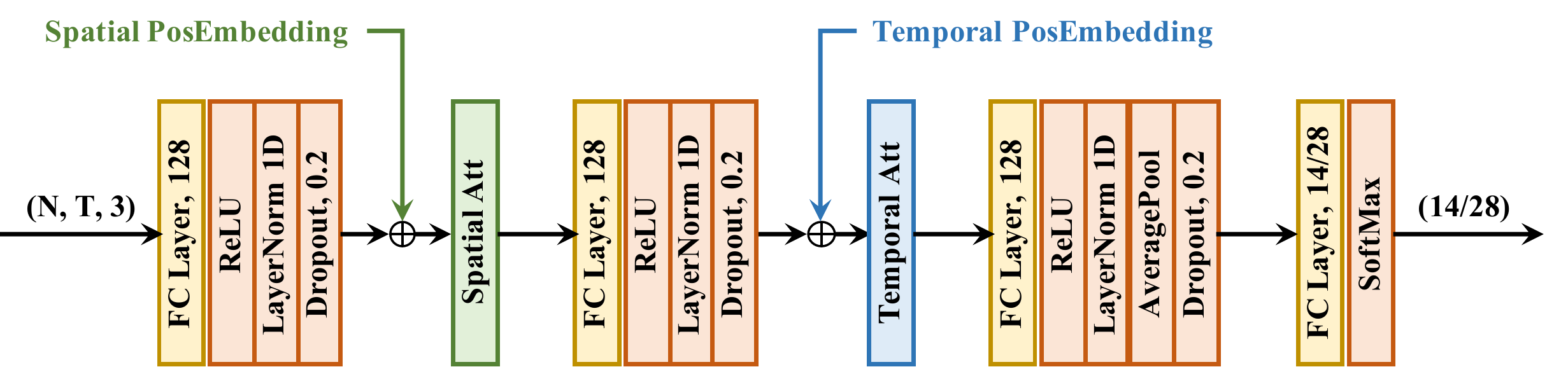}
\caption{The network architecture of the proposed DG-STA.}
\label{fig:arch}
\end{figure}

Our network structure is shown in Figure \ref{fig:arch}. We set the head number of the spatial and temporal attention models to 8. The dimension $d$ of the query, key and value vector is set to 32. Layer Normalization~\cite{ln} is utilized to normalize the intermediate outputs of our network. The input 3D coordinate of a hand joint is projected into an initial node feature of 128 dimension. It is then added with the corresponding spatial position embedding and fed into the spatial attention model, which produces a node feature of 256 dimension. This node feature is projected into a vector of 128 dimension which is added with the corresponding temporal position embedding. The temporal attention model takes it as the input and generates the final node feature. Finally, we average-pool the features of all nodes into a vector and feed it into a fully-connected layer for classification.

\subsection{Datasets and Settings}
\label{sec:settings}

We evaluate our method on the DHG-14/28 Dataset~\cite{de2016skeleton} and the SHREC'17 Track Dataset~\cite{de2017shrec}. Both datasets contain 2800 video sequences of 14 hand gestures which are performed in two configurations: using one single finger or the whole hand. The videos of the two datasets are captured by the Intel Realsense camera. The 3D coordinates of 22 hand joints in real-world space are provided per frame for network training and evaluation.

\textbf{Network Training.}
The proposed DG-STA is implemented based on the PyTorch platform. The Adam~\cite{kingma2014adam} optimizer with a learning rate of 0.001 is employed to train our model. The batch size is set to 32 and the dropout rate~\cite{srivastava2014dropout} is set to 0.2. We uniformly sample 8 frames from each video as the input. For fair comparison, we perform data augmentation by applying the same operations as proposed in~\cite{de2017shrec,nunez2018convolutional} including scaling, shifting, time interpolation and adding noise. We also subtract every skeleton sequence by the palm position of the first frame following Smedt~\etal~\cite{de2017shrec} for alignment.

\textbf{Evaluation Protocols.}
On the DHG-14/28 Dataset, models are evaluated by using the \emph{leave-one-subject-out cross-validation} strategy~\cite{de2016skeleton}. Specifically, we perform one experiment for each subject in this dataset. In each experiment, one subject is selected for testing and the remaining 19 subjects are used for training. The average accuracy of 14 gestures (without the single-finger configuration) or 28 gestures (with the single-finger configuration) over the 20 cross-validation folds are reported. For the SHREC'17 Track Dataset, we use the same data split as provided by~\cite{de2017shrec} and report the accuracy of both 14 and 28 gestures.

\subsection{Ablation Study}
\label{sec:abl_study}
Our proposed approach consists of three major components, including the fully-connected skeleton graph structure (FSG), the spatial-temporal attention model (STA) and the spatial-temporal position embedding (STE). We validate the effectiveness of these components in this section. The results are shown in Table~\ref{tbl:abl_std}.

\begin{table}[h]
\begin{center}
\begin{tabular}{|c|c|c|c|c|}
\hline
Setting & FSG+STA & FSG+GAT+STE & SSG+STA+STE & \textbf{DG-STA}\\
\hline\hline
14 Gestures~(D) & 84.3 & 90.8 & 89.8 & \textbf{91.9}\\
28 Gestures~(D) & 77.3 & 87.8 & 86.6 & \textbf{88.0}\\
\hline\hline
14 Gestures~(S) & 88.9 & 92.7 & 91.5 & \textbf{94.4}\\
28 Gestures~(S) & 80.1 & 86.2 & 87.7 & \textbf{90.7}\\
\hline
\end{tabular}
\end{center}
\caption{Ablation study of accuracy (\%) on the DHG-14/28 Dataset (D) and SHREC'17 Track Dataset (S). Our full model (DG-STA) achieves the best performance.}
\label{tbl:abl_std}
\end{table}

\textbf{Evaluation of Fully-Connected Graph Structure.} We compare the proposed FSG with the sparse skeleton graph structure (SSG) introduced by Yan~\etal~\cite{yan2018spatial}, where spatial edges are defined based on the natural connections of the hand joints and temporal edges connect the same joints between consecutive frames. We can see that our model significantly outperforms the one trained on SSG. This is because SSG may be sub-optimal for some hand gestures, while FSG has little constrains on the model so that it is able to learn action-specific graph structures.

\textbf{Evaluation of Spatial-Temporal Attention.} The proposed STA downgrades to Graph Attention (GAT)~\cite{velivckovic2017graph} if only one attention model is applied to the whole graph without distinguishing the spatial and temporal domains. We implement GAT by replacing the spatial and temporal attention models in our network with one attention model, and train it under the same setting of our model. We can observe that the STA-based model achieves better performance than the GAT-based model, which demonstrates the effectiveness of STA.

\textbf{Evaluation of Spatial-Temporal Position Embedding.} We validate the effectiveness of the proposed STE by training a variant of our method where STE is removed. We can see that our model outperforms the model without STE, which demonstrates the importance of the identity and temporal order information encoded by STE.


\subsection{Comparison with Previous Methods}
\label{sec:comparison}

We compare our method with the state-of-the-art methods on the DHG-14/28 Dataset~\cite{de2016skeleton} and the SHREC'17 Track Dataset~\cite{de2017shrec}, respectively. The compared state-of-the-art methods include traditional hand-crafted feature approaches~\cite{chen2017motion,oreifej2013hon4d,devanne20153,ohn2013joint,de2017dynamic,de2016skeleton,caputo2018comparing,boulahia2017dynamic}, deep learning based approaches~\cite{nunez2018convolutional,hou2018spatial,de2017shrec} and a graph-based method~\cite{yan2018spatial}. The results are shown in Tables~\ref{tbl:dhg} and \ref{tbl:SH17}. Note that for ST-GCN~\cite{yan2018spatial}, we implement it following the \emph{distance partitioning} setting and use a three-layer ST-GCN with 128 channels for fair comparison. We collect the results of other baseline methods from~\cite{hou2018spatial}.



\begin{table}[h]
\begin{center}
\begin{tabular}{|c|c|c|}
\hline
Method & 14 Gestures & 28 Gestures\\
\hline\hline
SoCJ+HoHD+HoWR~\cite{de2016skeleton} & 83.1 & 80.0\\
Chen~\etal~\cite{chen2017motion} & 84.7 & 80.3\\
CNN+LSTM~\cite{nunez2018convolutional} & 85.6 & 81.1\\
Res-TCN~\cite{hou2018spatial} & 86.9 & 83.6\\
STA-Res-TCN~\cite{hou2018spatial} & 89.2 & 85.0\\
ST-GCN~\cite{yan2018spatial} & 91.2 & 87.1\\
\hline\hline
\textbf{DG-STA (Ours)} & \textbf{91.9} & \textbf{88.0}\\
\hline
\end{tabular}
\end{center}
\caption{Comparisons of accuracy (\%) on DHG-14/28 Dataset.}
\label{tbl:dhg}
\end{table}

\textbf{Results on DHG-14/28 Dataset}. From Table~\ref{tbl:dhg}, we can see that our method achieves the state-of-the-arts performance under both 14-gesture and 28-gesture setting. Moreover, both our method and ST-GCN~\cite{yan2018spatial} outperform other methods which do not explicitly exploit structures and dynamics of hands, which demonstrates that these messages are important for skeleton-based hand gesture recognition.

\begin{table}[h]
\begin{center}
\begin{tabular}{|c|c|c|}
\hline
Method & 14 Gestures & 28 Gestures\\
\hline\hline
Oreifej~\etal~\cite{oreifej2013hon4d} & 78.5 & 74.0\\
Devanne~\etal~\cite{devanne20153} & 79.4 & 62.0\\
Classify Sequence by Key Frames~\cite{de2017shrec} & 82.9 & 71.9\\
Ohn-Bar~\etal~\cite{ohn2013joint} & 83.9 & 76.5\\
SoCJ+Direction+Rotation~\cite{de2017dynamic} & 86.9 & 84.2\\
SoCJ+HoHD+HoWR~\cite{de2016skeleton} & 88.2 & 81.9\\
Caputo~\etal~\cite{caputo2018comparing} & 89.5 & -\\
Boulahia~\etal~\cite{boulahia2017dynamic} & 90.5 & 80.5\\
Res-TCN~\cite{hou2018spatial} & 91.1 & 87.3\\
STA-Res-TCN~\cite{hou2018spatial} & 93.6 & \textbf{90.7}\\
ST-GCN~\cite{yan2018spatial} & 92.7 & 87.7\\
\hline\hline
\textbf{DG-STA (Ours)} & \textbf{94.4} & \textbf{90.7}\\
\hline
\end{tabular}
\end{center}
\caption{Comparisons of accuracy (\%) on SHREC'17 Track Dataset.}
\label{tbl:SH17}
\end{table}

\textbf{Results on SHREC'17 Track Dataset}. Different from the DHG-14/28 Dataset where videos are cropped by human-labeled beginnings and ends of the gestures~\cite{de2016skeleton}, the SHREC'17 Track Dataset provides raw captured video sequences with noisy frames, and hence is more challenging. We can see that our method achieves the state-of-the-arts performance under the 14-gesture setting, and obtains comparable performance with STA-Res-TCN~\cite{hou2018spatial} under the 28-gesture setting. In addition, we can observe that our method and ST-GCN~\cite{yan2018spatial} outperform all other methods which do not explicitly exploit structures and dynamics of hands.

\section{Conclusions}

In this paper, we proposed a graph-based spatial-temporal attention method for skeleton-based hand-gesture recognition. It utilizes two attention models in the spatial and temporal domains of the fully-connected hand skeleton graph to learn edge weights and extract spatial and temporal information for hand gesture recognition. Extensive experiments demonstrate
the effectiveness of our framework. Our proposed method provides a general framework that can be further used for other tasks aiming to learn spatial and temporal information from graph-based data, \eg, skeleton-based human action recognition.

\section{Acknowledgments}

This work was funded partly by ARO-MURI-68985NSMUR and NSF 1763523, 1747778, 1733843, 1703883 to Dimitris N. Metaxas.


\end{document}